\documentclass[manuscript, nonacm]{acmart}
\AtBeginDocument{%
  \providecommand\BibTeX{{%
    \normalfont B\kern-0.5em{\scshape i\kern-0.25em b}\kern-0.8em\TeX}}}

\setcopyright{rightsretained}
\copyrightyear{2024}
\acmYear{2024}
\acmDOI{XXXXXXX.XXXXXXX}

\acmJournal{JACM}
\acmVolume{37}
\acmNumber{4}
\acmArticle{111}
\acmMonth{8}

\newcommand{\Sim}{ \mathcal{S} }
\newcommand{\Render}{ \mathcal{R} }
\newcommand{\Obj}{ \mathcal{O} }
\newcommand{\Q}{ \mathcal{Q} }
\newcommand{\V}{ \mathcal{V} }
\newcommand{\T}{ \mathcal{T} }
\newcommand{\s}{ \mathbf{s} }
\newcommand{\p}{ \mathbf{p} }
\newcommand{\h} {\mathbf{h}}
\newcommand{\f} {\mathbf{f}}
\newcommand{\F} {\mathbf{F}}
\newcommand{\W} {\mathbf{W}}
\newcommand{\I} {\mathbf{I}}
\newcommand{\real} {\mathbb{R}}

\begin{document}

\title{Differentiable Rendering as a Way to Program Cable-Driven Soft Robots}


\author{Kasra Arnavaz}
\email{kasra@di.ku.dk}
\orcid{0000-0001-9228-9064}

\author{Kenny Erleben}
\email{kenny@di.ku.dk}
\orcid{0000-0001-6808-4747}
\affiliation{%
  \institution{Department of Computer Science, University of Copenhagen}
  \streetaddress{Universitetsparken 1}
  \city{Copenhagen}
  \country{Denmark}
  \postcode{2100}
}

\renewcommand{\shortauthors}{Arnavaz and Erleben}

\begin{abstract}
    Soft robots have gained increased popularity in recent years due to their adaptability and compliance. In this paper, we use a digital twin model of cable-driven soft robots to learn control parameters in simulation. In doing so, we take advantage of differentiable rendering as a way to instruct robots to complete tasks such as point reach, gripping an object, and obstacle avoidance. This approach simplifies the mathematical description of such complicated tasks and removes the need for landmark points and their tracking. Our experiments demonstrate the applicability of our method.
\end{abstract}



\received{11 April 2024}

\maketitle

\section{Introduction}
While rigid robots are suitable for industrial fabrication due to their precision, they are unfitting for home applications where contact safety is required \cite{lipson.14}. Inspired by nature, soft robots are more adaptable and flexible, making them suitable for use cases with little prior knowledge about the environment \cite{kim.ea13}. As a result, they have been used in mechanical gripping \cite{shintake.ea18}, exploration \cite{aracri.ea21}, and healthcare \cite{zhong.ea20, cianchetti.ea18}. This very flexibility, however, makes modeling and control of soft robots difficult, due to their high degrees of freedom \cite{dubied.ea22}. Soft robots are underactuated, and it is costly to predict the effect of changing a control parameter through trial and error.  

Previous work has addressed this challenge by applying optimization of model-based control, i.e. having a physical simulation model to help prevent the consequences of changing the control signals \cite{coevoet.ea17}. This is promising as a computer algorithm now can numerically explore the space of possible control solutions. Having said that, there are two open questions: 1) how to reduce the reality gap and 2) how to program the robot, i.e. how one would mathematically set up goals to be achieved. In \cite{arnavaz.ea23, murthy.ea21}, rendering and computer vision combined with differentiability have shown great strength in addressing the first challenge through the optimization of material parameters to ensure the simulated deformations are as close as possible to the ones observed in the real world.

In this work, we are inspired by how differentiability and rendering have created an elegant solution to simulator calibration, and we explore this paradigm as a means to more easily program complex tasks for soft robots. Such tasks are often expressed as point constraints or point-tracking objectives \cite{largilliere.ea15, duriez.13}. One needs to describe these functions to achieve desired behaviors. It is then up to manual trial and error to add more constraints to the setup. Take, for example, a gripping task, where a soft robot needs to conform to the surface of a geometric shape to maximize contact area resulting in a firm grasp. One would have to place multiple-point samples manually on these surfaces to get an objective that yields the desired outcome. Or the case where the robot needs to avoid dangerous areas in space, where other objects or fabrication processes (e.g. a water jet cutter) could physically damage it. One solution is to put dummy shapes in a simulation to force a soft robot away from those no-touch zones, but that would bias the solution since one relies on contact forces between a robot and a dummy shape to stay away.

We recast such scenarios into rendering interpretations such that these tasks are redefined through depth images thus eliminating the need to define point correspondences to solve tasks. In summary, our contributions to this paper are

\begin{itemize}
    \item formulating grasping and avoidance tasks for curved surfaces with differentiable rendering,
    \item proposing a simple mathematical description for maximal contact area for a gripping task,
    \item and instructing the soft robot to avoid certain zones without contact forces.
\end{itemize}

\section{Related Works}
\paragraph{Modeling} Soft robots are infinite-dimensional in their degrees of freedom, making modeling them a trade-off between tractability and accuracy \cite{della.ea23}. Analytical models tend to impose restrictive assumptions on the dynamics of the robot \cite{schegg.duriez.22, wang.ea13, renda.ea18}, limiting the generality of their applications. Numerical methods have proven to be more precise in practice \cite{schegg.duriez.22} and can tackle shapes of higher complexity at a higher cost of computation. Among them, the Finite Element Method (FEM) has shown success in describing nonlinear dynamics as well as contact interactions with objects \cite{bartlett.ea15, courtecuisse.ea14}. 

\paragraph{Simulation} Several engines have been developed based on various modeling techniques. Since differentiability is a core part of our pipeline, we focus our attention on differentiable simulators suited for soft deformable bodies. To name a few, DiffAqua \cite{ma.ea21}, ChainQueen \cite{hu.ea19}, and Warp \cite{warp.22} have been used for soft deformable bodies. While we use Warp in this work, our method is not limited to this choice and can be substituted with any differentiable simulator. Our work is closest to \cite{arnavaz.ea23, murthy.ea21}, where differentiability has been utilized to calibrate the physical properties of soft bodies. We utilize differentiable rendering as a language to express soft robotics tasks and tackle control problems 

\paragraph{Control} Actuation of soft robots can be divided into pneumatic- and control-based methods. The latter relies on air pressure change and the former on cable length change to control the behavior of soft robots. In this paper, we focus on cable-driven robots. Model-based control has been extensively surveyed in \cite{della.ea23}. In particular, we use the FEM to iteratively simulate the behavior of the soft robot and use physical modeling of the cables by damped springs to obtain cable forces. While model-free approaches are getting more popular \cite{tiboni.ea23}, they require iterative time-consuming training to find control strategies \cite{wu.ea23}.

\section{Method}
\begin{figure}[ht]
    \centering
    \includegraphics[width=0.9\linewidth]{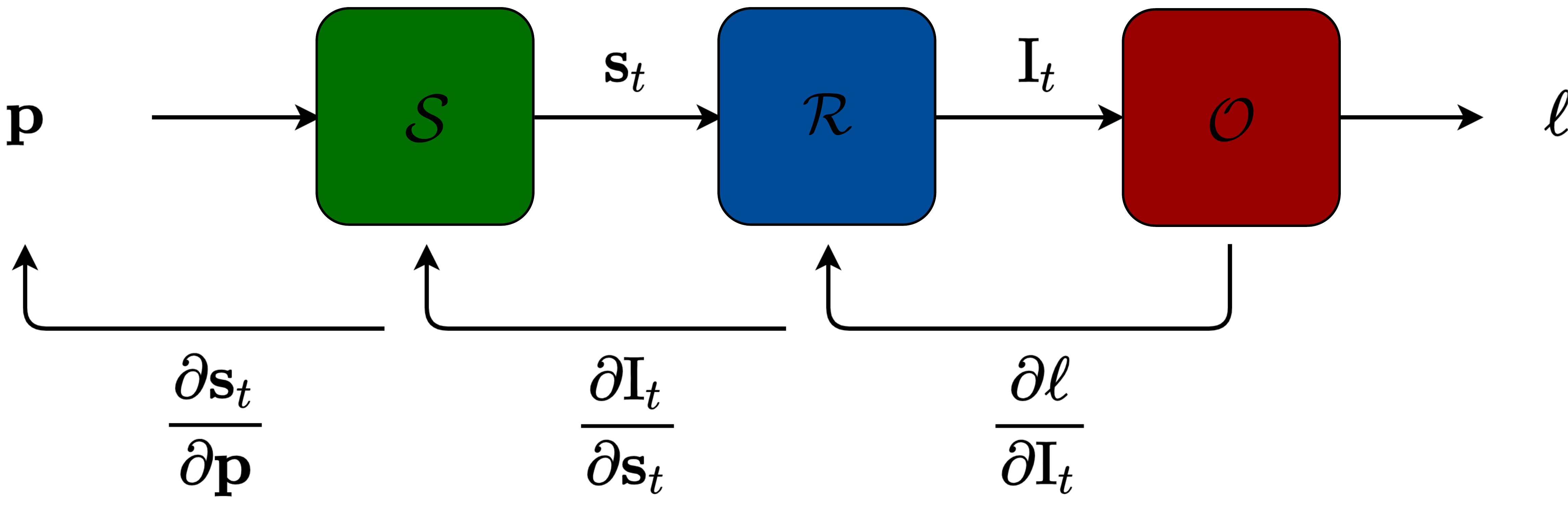}
    \caption{The data flow of our differentiable pipeline. The simulation function $\Sim$ produces a state $\s_t$ using control parameters $\p$. The state is then projected to an image $\I_t$, which depending on the objective $\Obj$ is converted to a scalar loss $\ell$. The chain rule can then be applied to compute the derivative of $\ell$ w.r.t. control parameters $\p$.}
    \Description[Data flow visualization]{Illustration showing how the differentiable pipeline works.}
    \label{fig:pipeline}
\end{figure}

We argue that one could solve tasks such as gripping and obstacle avoidance by modeling the contact surface between the robot and the objects in the scene. We achieve this goal by rendering this surface by viewing the world from the object's perspective, i.e. we place cameras inside the object and render the object's surface and the robot. With those depth images, we propose a fully differentiable pipeline (Figure \ref{fig:pipeline}) to learn the control parameters of cable-driven soft robots.

Our approach has three main functions, namely simulation, rendering, and objective functions. The simulation function takes the control parameters $\textbf{p}$ as input and outputs the state of the scene after a duration of $t$ seconds, i.e. $\s_t = \Sim(\textbf{p})$. The state contains the position and velocity of all mesh nodes in the scene. To get $\s_t$, one needs to compute the cable forces as well as elastic forces and gravitational force iteratively---which we cover in Section \ref{sec:force}. The rendering function takes the state of the three-dimensional scene and converts it to two-dimensional depth images for every mesh in the scene, i.e. $\I_t = \Render(\s_t)$. The intensity of these grayscale images conveys how far the mesh is from the camera, with darker pixels denoting closeness to the camera and brighter pixels denoting further distance to the camera. Finally, the objective function takes these depth images and outputs a scalar loss, i.e. $\ell = \Obj(\I_t)$.

\subsection{Cable Force}
\label{sec:force}
To simulate the behavior of the cable-driven robot, we need to model how cable forces are computed. Once the forces are obtained, the simulation engine does numerical integration to calculate the resulting change in position and velocity.

\begin{figure}[ht]
    \centering
    \includegraphics[width=0.9\linewidth]{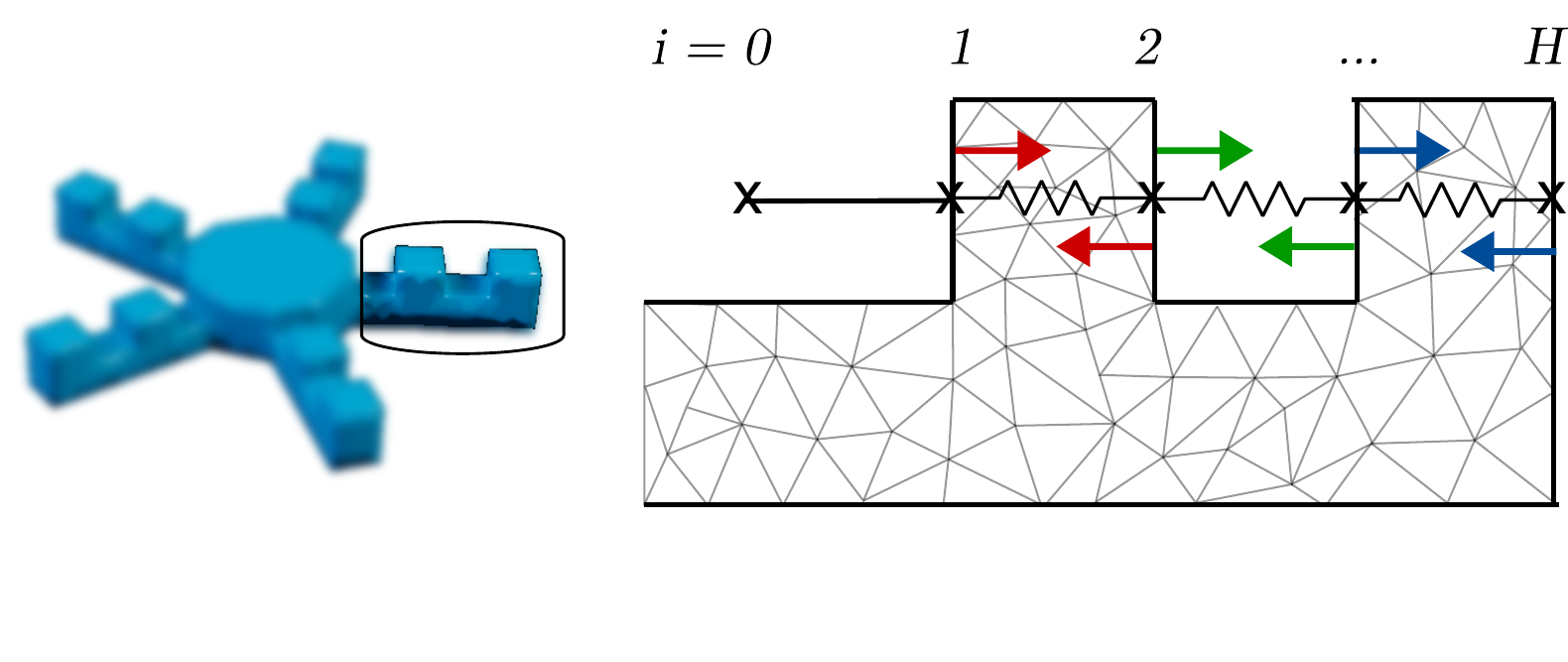}
    \caption{\textbf{Left:} A soft robot design whose fingers are curled by a cable running through them. \textbf{Right:} A depiction of how cable forces are computed for one finger. Cables in each segment are modeled with identical springs with damping to obtain forces at via points.  Barycentric coordinates are then used to map these forces to nodal points.}
            \label{fig:force}
    \Description[Cable forces]{Illustration showing how the cable forces are computed and coupled to the soft robot.}
\end{figure}

Consider the robotic finger shown in Figure \ref{fig:force}, which has one cable running through it. We are interested in computing the forces at each point where the cable enters/leaves the robot, also known as via points. We will model the cable forces by multiple identical springs at each interval. For every force at each via point, there is an equal but opposite force in the next via point where the spring ends---shown in Figure \ref{fig:force} by arrows of the same color. Given that we have $H$ via points for a cable, the forces corresponding via point $i$ are computed as 
\begin{equation}
    \f_i =
\left\{
	\begin{array}{ll}
		pk(\h_2 - \h_1) - b\dot \h_1 & \mbox{if } i = 1, \\
		-\f_{H-1} - b\dot \h_H & \mbox{if } i = H, \\
            -\f_{i-1} + \text{p}k(\h_{i+1} - \h_{i}) - b\dot \h_i & \text{otherwise,}
	\end{array}
\right.
\end{equation}
where $\h_i, \dot \h_i, \f_i \in \real^3$ are the position, velocity, and force of via point $i$ respectively, $k$ is the spring stiffness, $b$ is the damping coefficient and $p \in \real$ is the control parameter we would like to determine to make our robot achieve a particular task. If we denote the position of the tip of the cable by $\h_0$, then $p$, also referred to as the pull ratio, is given by
\begin{equation}
    \label{eq:p}
    p = \frac{||\h_1 - \h_0||_2}{||\h_H - \h_1||_2} ~.
\end{equation}
If the cable is not pulled at all, then $\h_0 = \h_1$ or equivalently $p=0$, and since $\dot \h_i=\vec 0$, then all cable forces at via points would be zero ($\f_i = \vec 0$). If a robot has $C$ cables, all pull ratios are stored in a vector denoted by $\textbf{p} \in \real^C$.

Thus far, we have computed the cable forces at the position of via points, but our simulation requires having the cable forces at each node. To perform this mapping, we will be using the barycentric coordinates of $\h_i$ inside the tetrahedron that contains it. If our mesh has $N$ nodes, then we will be getting a matrix of barycentric coordinates $\W \in \real^{H \times N}$, which sums to one in each row by definition of barycentric coordinates.  Let us collect forces at every via point in a column vector defined as $\f = \left(\f_1, \f_2, ..., \f_H\right)^T$. Similarly, if we denote the force at node $j$ with $\F_j$, we have $\F = \left(\F_1, \F_2, ..., \F_N\right)^T$. We would then have the relationship $\f = \W\F$. Given that $N \ll H$, $\W^T\W$ is invertible and we have
\begin{equation}
    \F = (\W^T\W)^{-1}\W^T\f.
\end{equation}
These cable forces are going to be added to the elastic forces as an external force, i.e.
\begin{equation*}
    \F_{\text{total}} = \sum_{c \in \mathcal{C}}^{} \F^{c} + \F_{\text{elastic}} + \F_{\text{gravity}},
\end{equation*}
where $\F^c$ denotes the nodal force matrix of cable $c$ and $\mathcal{C}$ is the set of cables in a given soft robot. Note that $\F_{\text{elastic}}$ depends on the type of material the soft robot is made of, which can be specified with Young's modulus, Poisson's ratio, density, and damping factor.

\subsection{Robot Tasks}
Robots in isolation cannot achieve many meaningful tasks. In our experiments in Section \ref{sec:exp}, we are interested in the interaction of the soft robot with other objects. In particular, we are interested in programming the robot to a) grip an object with maximal contact and b) avoid obstacles while reaching a point in space. The key ingredient to solving both these tasks is having a measure of distance between the robot and the object of interest. Then, the obstacle avoidance task would translate to learning control parameters to maximize this distance while gripping corresponds to minimizing it.

\begin{figure}[ht]
    \centering
    \includegraphics[width=0.9\linewidth]{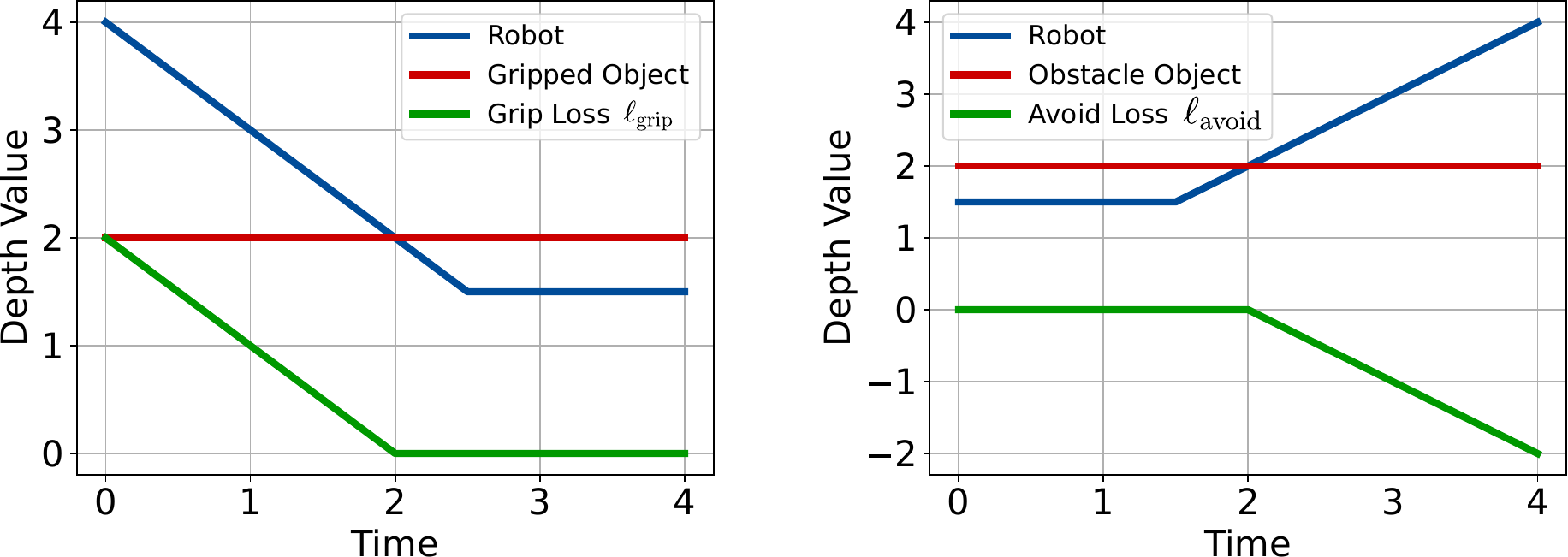}
    \caption{Didactic plots for depicting loss values given a robot depth map. \textbf{Left:} When gripping, the robot gets closer to the object over time and might penetrate it depending on the collision resolution. \textbf{Right:} The robot is assumed to be penetrating the obstacle initially and gets further away from it with time. Notice, how lower loss values would translate to gripping and avoiding in each case. In both cases, the object does not move, and as a result, its depth value remains constant.}
    \label{fig:didactic}
    \Description[Loss plots]{Illustration of how we wish loss plots to appear.}
\end{figure}

Let us denote the robot by $r$, the object we want to grip or the point to reach by $s$, and the set of obstacle objects in the scene by $\Q$. To obtain such a distance measure, we will place the camera at various angles inside the object of interest, i.e. the obstacle we want to avoid or the object we would like to grip. The set of camera views for each object is defined by $\V$, and the set of time frames where an image has been recorded is defined by $\T$. Finally, the depth image of the robot $r$ and the object $s$ at time $t \in \T$ with camera view $v \in \V$ with $\I^{r}_{t}(v)$ and $\I^{s}_{t}(v)$, respectively. Then the distance measure is defined by  
\begin{equation}
    \Delta \I_v(r,s,t) \equiv \text{max}(\I^{r}_{t}(v) - \I^{s}_{t}(v), 0),
    \label{eq:distance}
\end{equation}
where $\text{max}(\cdot, 0)$ sets the distance to zero when the robot penetrates the object. We can then write the loss of the gripping task by robot $r$ on object $s$ as
\begin{equation}
\label{eq:grip}
    \ell_{\text{grip}}(r,s) \equiv \frac{1}{|\T||\V|}\sum_{v \in \V}^{}\sum_{t \in \T}^{} \Delta \I_{v}(r,s,t).
\end{equation}
Similarly, the loss corresponding to the avoidance of obstacle $q$ is defined as
\begin{equation}
\label{eq:avoid}
    \ell_{\text{avoid}}(r,q) \equiv -\frac{1}{|\T||\V|}\sum_{v \in \V}^{}\sum_{t \in \T}^{} \Delta \I_{v}(r,q,t).
\end{equation}
Both tasks can be achieved simultaneously with a multi-object loss as
\begin{equation}
\label{eq:proa}
    \ell(r, s) \equiv \alpha\ell_{\text{grip}}(r,s) + \frac{(1-\alpha)}{|\Q|}\sum_{q \in \Q}^{} \ell_{\text{avoid}}(r,q),
\end{equation}
where the first part encourages reaching the point $s$, the second part penalizes getting close to the obstacles, and $\alpha \in [0, 1]$ is the weighting parameter between the two parts. Figure \ref{fig:didactic} illustrates the structure of gripping and avoidance losses with an example.

\section{Experiments}
\label{sec:exp}

We have conducted four experiments to demonstrate the effectiveness and simplicity of our approach. We have viewed the robot-object interactions as attraction (e.g. point reach and gripping) and repulsion (e.g. obstacle avoidance). Differentiable rendering has proven to be a potent tool in modeling both scenarios and learning the control parameters with gradient-based methods. We have used Pytorch3D \cite{johnson.ea20} as our differentiable renderer and Gradient Descent as our optimizer. Initially, the pull ratio introduced in Equation (\ref{eq:p}) is set to zero for all cables in all experiments and then updated according to the gradients. Furthermore, pull ratios are clipped to zero, if their value is negative. In the following, we describe each experiment. Table \ref{tab:common} contains the parameters that were common in all experiments, and Table \ref{tab:unique} shows the custom parameters for each experiment.

\begin{table}[ht]
    \caption{Common parameters in all experiments}
    \label{tab:common}
    \setlength{\tabcolsep}{45pt}%
    \begin{tabular}{@{}ll}
        \toprule
         Parameter & Value  \\
        \midrule
           Young's modulus &   149 kPa \\
           Poisson's ratio &   0.40 \\
            Damping factor &   0.40 \\
           Density &   1080 kg/m³ \\
           Cable stiffness & 100 N/m   \\
           Cable damping coefficient & 0.01 kg/s  \\
           Simulation time step  & $5 \times 10^{-5}$ s\\
        \bottomrule
    \end{tabular}
\end{table}

\begin{table}[ht]
    \caption{Experiment-specific parameters}
    \label{tab:unique}
    \setlength\aboverulesep{0pt}
    \setlength\belowrulesep{0pt}
        \begin{tabular}{l|ccccc}
        \toprule
        Experiment & Cables & Vertices & Tetrahedra & Duration & Learning rate\\
        \midrule
           Reach & 3 & 2830 & 10748 & 2 s & 0.1\\
           Avoidance & 3 & 2830 & 10748 & 2 s & 1.0\\
            Cylinder & 5 & 2559 & 9407 & 3 s  & $10^{-5}$\\
           Egg & 5 & 764 & 1753 & 2 s & $10^{-3}$\\
        \bottomrule
    \end{tabular}
\end{table}

\begin{figure}[t]
    \centering
    \includegraphics[width=0.9\linewidth]{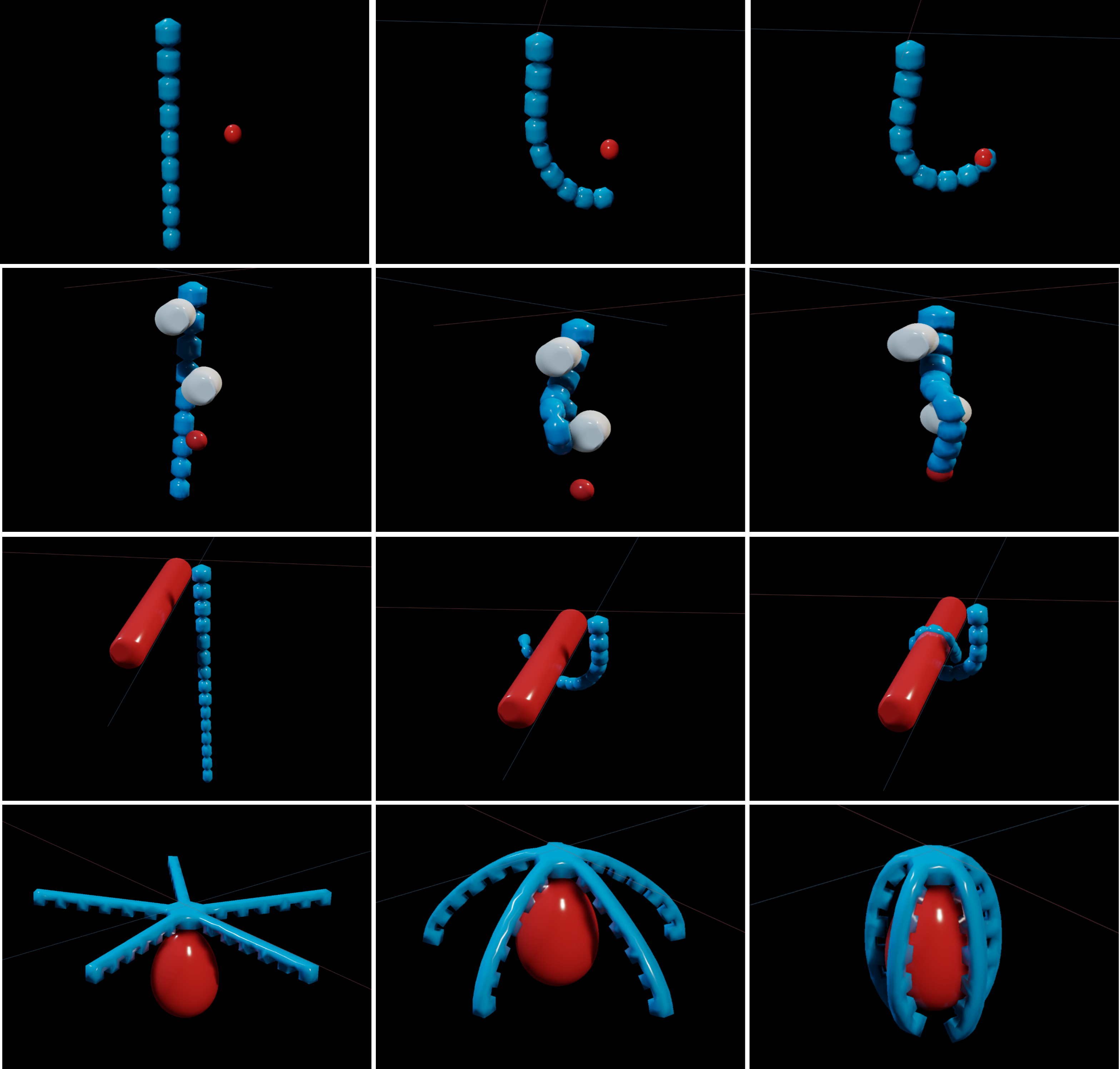}
    \caption{Snapshots of the scene showing the progression of tasks. \textbf{First row:} Reach Experiment at 0, 0.2, 2 seconds. \textbf{Second row:} Avoidance Experiment at 0, 0.2, 2 seconds. \textbf{Third row:} Cylinder Experiment at 0, 0.5, 3 seconds. \textbf{Forth row:} Egg Experiment at 0, 0.25, 2 seconds.}
    \label{fig:omni}
    \Description[Examples of tasks]{Illustration showing how different tasks are solved.}
\end{figure}

\begin{figure}[t]
    \centering
    \includegraphics[width=0.9\linewidth]{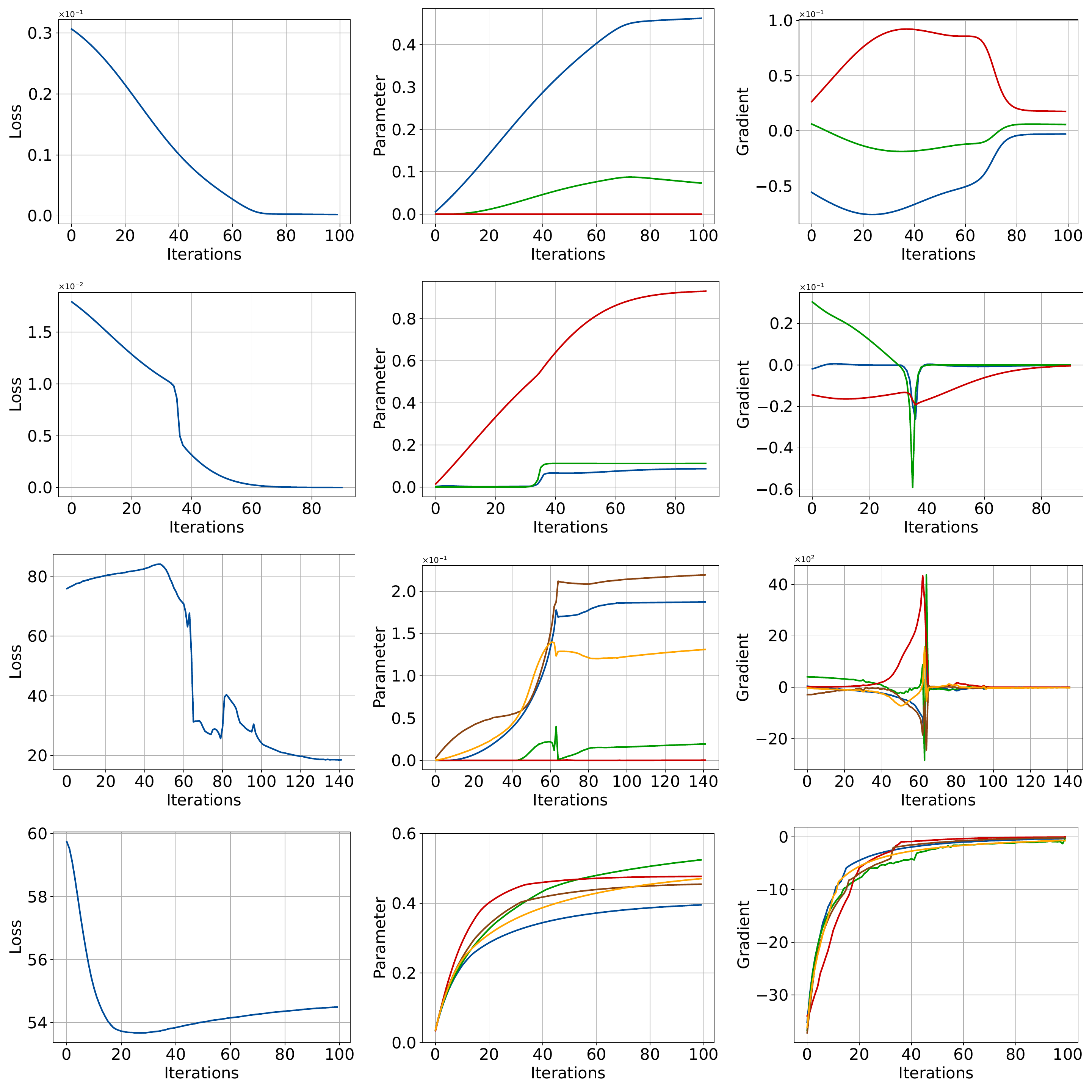}
    \caption{Convergence plots for all experiments. \textbf{First row:} Reach Experiment. \textbf{Second row:} Avoidance Experiment. \textbf{Third row:} Cylinder Experiment. \textbf{Fourth row:} Egg Experiment. The columns show each cable's loss, parameter values, and gradients for each cable.}
    \label{fig:convergence}
    \Description[Convergence plots]{Gallery showing all convergence plots for all tasks that we solved.}
\end{figure}

\paragraph{Reach Experiment} The goal in this experiment is to get the robot to reach a point in 3D space (Figure \ref{fig:omni} first row). The red ball denotes the point to be reached. There are no contact forces between the robot and the ball and as a result, it does not exist from the robot's standpoint. Six cameras have been placed inside the ball to compute the distance measure in Equation (\ref{eq:distance}). The robot has three cables and hence three pull ratio parameters to be optimized. The loss, pull ratio values, and gradients are shown in Figure \ref{fig:convergence} first row.

\paragraph{Avoidance Experiment} Building on top of the Reach Experiment, cylindrical obstacles are now inserted on the robot's path to reach the point (Figures \ref{fig:omni} and \ref{fig:convergence} second row). Six cameras are placed inside all three objects looking in every direction. None of the objects in this scene produce contact forces (not the ball nor the obstacles). The obstacles are therefore only there as areas where the robot is not supposed to go, and there are no collision forces that prevent it from touching the obstacles. The robot is discouraged from touching the obstacles by the increase in loss and is encouraged to reach the ball by the reduction in loss as described in Equation (\ref{eq:proa}). After some trial and error, the weight parameter $\alpha$ is set to $0.7$.

\begin{figure}[ht]
    \centering
    \includegraphics[width=0.9\linewidth]{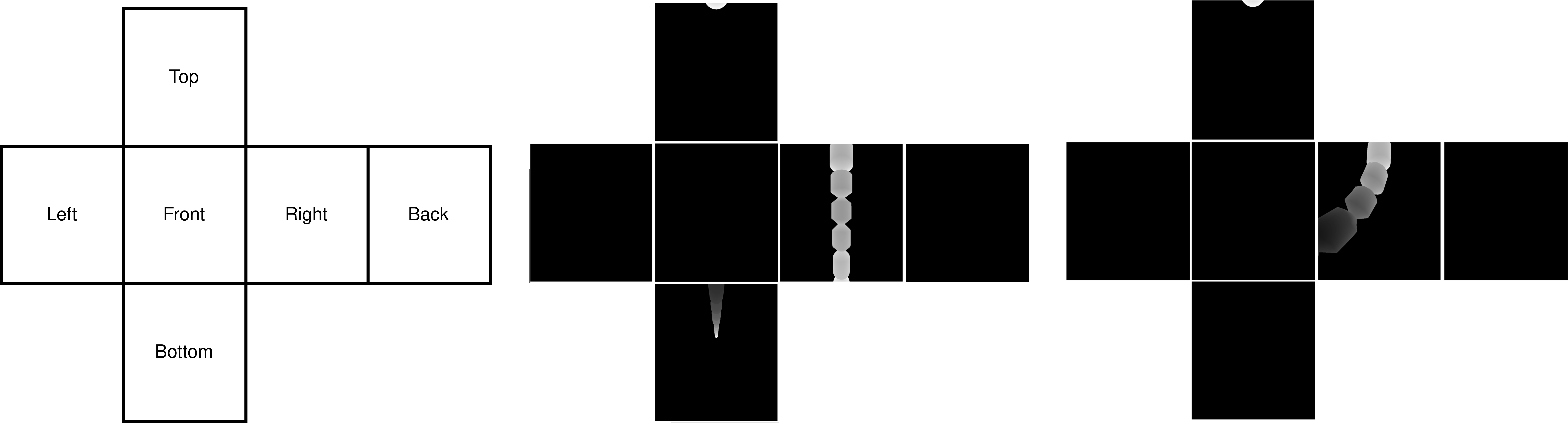}
    \caption{The distance measure introduced in Equation (\ref{eq:distance}) for Cylinder Experiment. Brighter values denote further distance between the robot and the object and vice versa. \textbf{Left:} The cube map for all six camera views. \textbf{Middle:} The distance measure before optimization of control parameters ($\p = \vec{0}$). \textbf{Right:} The distance measure as a result of optimized control parameters. }
    \label{fig:loss}
    \Description[Loss]{Plot showing distance measure as a result of optimized control parameters.}
\end{figure}

\paragraph{Cylinder Experiment} The trunk robot in this task is designed to be longer compared to previous tasks to showcase the maximal contact behavior in gripping around the cylinder shaft. Figure \ref{fig:loss} shows the distance measure (\ref{eq:distance}) before and after optimization for all six cameras placed inside the cylinder. Notice how the distance is smaller after optimization when the cylinder has been gripped. By lowering the loss in Equation (\ref{eq:grip}) and enabling the collision detection in the simulation engine, the robot curls around the cylinder shaft (Figures \ref{fig:omni} and \ref{fig:convergence} third row).

\paragraph{Egg Experiment} To demonstrate that this approach applies to other designs of soft robots, we have made a starfish gripper with five fingers with one cable running through each finger. The task is to have a maximal contact area with an eggshell object (Figures \ref{fig:omni} and \ref{fig:convergence} fourth row). The experimental setup is similar to that of the Cylinder Experiment. 

\section{Conclusion}
This paper presents a novel approach to programming robots using differentiable rendering. The strength of our method lies in the simplicity of formulating complex robotics tasks such as mechanical gripping of objects and obstacle avoidance, both of which can be expressed with depth images obtained from the interior view of these objects. This eliminates the need to come up with point correspondences and track these landmarks, a complex computer vision task with real-world data. Our experiments demonstrate the effectiveness of this programming paradigm for soft robots. A limitation of our work is that learned control parameters apply only to those specific predefined environments. It is a future work to combine this framework with reward functions and reinforcement learning to learn control \emph{policies}, which would be more robust to changes in the environment.

\bibliographystyle{ACM-Reference-Format}
\bibliography{references}

\end{document}